%% file: main_en.tex
\documentclass{article}

\PassOptionsToPackage{numbers}{natbib}
\usepackage[main, final]{neurips_2026}

\usepackage{iftex}
\ifPDFTeX
  \usepackage[utf8]{inputenc} 
  \usepackage[T1]{fontenc}    
\fi
\ifXeTeX
  \usepackage{fontspec}
  \usepackage{xeCJK}
  \setCJKsansfont{Songti SC}
  \setCJKmonofont{Songti SC}
\fi
\usepackage{hyperref}
\usepackage{url}
\usepackage{graphicx}
\usepackage{booktabs}
\usepackage{amsfonts}
\usepackage{nicefrac}
\usepackage{microtype}
\usepackage{xcolor}

\title{APPSolver: Adaptive Patch Partitioning for Point-Wise Ship Flow Prediction on Unstructured Meshes}

\author{
Wenhua Huo \quad Fenglei Han\thanks{Corresponding author.} \quad Wangyuan Zhao \quad Xiao Peng \\
Chunhui Wang \quad Jialin Wu \quad Jiayi Han \\
College of Shipbuilding Engineering, Harbin Engineering University \\
\texttt{fenglei\_han@hrbeu.edu.cn}
}

\begin{document}

\maketitle

\begin{abstract}
Predicting high-fidelity flow fields around ship hulls is central to hydrodynamic performance analysis, yet traditional CFD solvers remain computationally prohibitive for rapid design exploration across multiple vessel types and operating conditions.  Existing deep-learning surrogates often interpolate unstructured CFD meshes onto regular grids, which can blur near-field resolution, while direct point-level global attention on large unstructured meshes can be computationally expensive. In this paper, we introduce APPSolver, a neural surrogate that learns next-step point-wise flow-field prediction directly on the unstructured free-surface ship CFD mesh without re-gridding. APPSolver is built around Adaptive Patch Partitioning (APP), a quadtree-based strategy that groups unstructured points into local spatial patches according to the non-uniform density inherent in ship CFD meshes: fine near the hull and wake, coarse in the far field. The resulting patch tokens are processed by a Transformer backbone together with optional condition token that encode ship geometry and condition parameters via a frozen large language model, forming a unified token-space model for one-step temporal advancement of velocity and pressure fields. On the ShipBench dataset, APP-Transformer exhibits an accuracy--efficiency trade-off, incurring higher prediction errors than the strongest learned baselines while requiring substantially lower forward computation.  Ablation studies quantify the compression-fidelity trade-off of APP and demonstrate that condition token provide preliminary gains in some leave-one-hull-out settings. Code is available at \url{https://github.com/wenhuahuo/APPSolver}.
\end{abstract}

\section{Introduction}

High-fidelity prediction of flow fields around ship hulls is critical for hydrodynamic performance analysis, yet traditional CFD methods remain computationally expensive for rapid design exploration across multiple hull types and operating conditions \citep{stern2013computational,visonneau2022computational}. Deep learning surrogate modeling has been widely used to accelerate flow field prediction \citep{wang2024recent}, with temporal advancement operators offering a natural formulation for unsteady problems. However, ship CFD meshes are highly non-uniform---finely refined near the hull and wake, coarser in the far field \citep{wackers2010adaptive}---and existing approaches face two key challenges when applied to such data.

First, many surrogate models interpolate flow fields onto regular grids for compatibility with CNNs \cite{khanal2025comparison}, Vision Transformers \citep{miotto2023flow}, or Fourier Neural Operators \citep{li2020fno}. While regular grid representations facilitate model processing, the interpolation process may diminish the local refinement advantages of unstructured meshes in near-wall and wake regions. Neural operators such as GNOT \citep{hao2023gnot}, Transolver \citep{wu2024transolver}, and UPT \citep{alkin2024upt} handle irregular meshes directly, but they are not suitable for handling extreme spatial density inhomogeneities in the free-surface sets of ships. Second, incorporating geometric and operating-condition information typically requires task-specific modules \citep{alkin2024upt,pfaff2020learning,hao2023gnot}, limiting extensibility to new configurations.

To address these challenges, we propose Adaptive Patch Partitioning (APP), a quadtree-based strategy that groups unstructured mesh points into local patches according to their spatial density---finer near the hull and wake, coarser in the far field. APP converts the non-uniform point set into compact patch tokens compatible with standard Transformers while preserving the original mesh's spatial organization. Combined with condition token encoded from ship geometry and condition parameters via a frozen language model, APPSolver performs next-step point-wise flow field prediction directly on the original unstructured CFD mesh. The APP method also provides a unified and flexible implementation for incorporating ship geometric parameters and condition configurations: geometric and condition parameters are mapped into condition token through independent encoders and concatenated with flow field patch tokens, enabling the model to jointly model local flow field evolution, global hull geometry influence, and operating condition effects in a unified token space.

Our main contributions are summarized as follows:

\begin{itemize}
    \item We formulate ship flow field prediction as a one-step temporal advancement surrogate modeling task on the unstructured free-surface ship CFD mesh. Unlike regular grid flow field prediction or steady-state neural operators, we directly learn flow field evolution mappings on the original unstructured mesh, thereby better preserving high-resolution information in the ship near-field region.

    \item We propose the Adaptive Patch Partitioning method, which adaptively partitions large-scale unstructured meshes into local patches and encodes each patch into a standard Transformer token, reducing computational complexity while preserving the spatial distribution characteristics of the original CFD mesh.
    
    \item We propose a unified condition-aware token fusion framework. Based on the standard patch tokens generated by APP, ship geometric parameters and operating condition parameters can be encoded as condition token and uniformly fused with patch tokens, enabling the model to flexibly generalize across different ship types and conditions.
\end{itemize}

\section{Related work}

We review related work along two directions below. Additional discussion on data-driven ship hydrodynamics surrogates and token-based representations for physical fields is provided in Appendix~\ref{app:related_work}.

\subsection{Regular-grid flow prediction and time rollout models}

Many flow field prediction models represent CFD data as regular grids or image tensors \citep{guo2016convolutional} and use convolutional networks \citep{ribeiro2020deepcfd}, U-Net \citep{luo2023cfdbench}, Vision Transformer \citep{miotto2023flow}, or Fourier Neural Operator \citep{li2020fno} for learning. Early work has demonstrated the effectiveness of convolutional neural networks in two-dimensional flow field prediction and surrogate modeling. FNO further approaches from the operator learning perspective, learning function-space-to-function-space mappings through frequency-domain convolution on regular grids, and has demonstrated high inference efficiency across multiple PDE benchmarks. In benchmarks closer to engineering CFD, regular grid representations have also been used to construct cross-geometry, cross-boundary-condition flow field prediction tasks, supporting autoregressive time rollouts \citep{luo2023cfdbench,jiang2023transcfd}.

The advantage of regular grid methods lies in uniform data shapes, high tensor computational efficiency, and the ability to directly leverage mature image modeling architectures. However, for ship CFD scenarios, the original mesh is typically locally refined around the hull near-field, free surface, and wake regions, while the far-field region is relatively sparse. Interpolating such non-uniform meshes onto regular grids may diminish the original resolution advantages in near-wall and wake regions while introducing redundant sampling in the far field. We do not deny the effectiveness of regular grid models in standard flow field prediction, but further focuses on how to learn one-step temporal advancement point-wise flow field prediction while preserving the spatial organization of ship unstructured CFD meshes.

\subsection{Operator learning on irregular meshes and complex geometries}

Beyond regular grid methods, another line of work directly targets physical field learning on unstructured meshes, point clouds, or complex geometries. MeshGraphNets \citep{pfaff2020learning} organizes mesh nodes and adjacency relationships into graph structures, learning the temporal evolution of physical systems through message passing, demonstrating the ability to model variable topologies and complex boundary conditions. Subsequent graph neural network methods have also been applied to fluid prediction tasks on unstructured meshes, showing that explicitly utilizing mesh connectivity helps model local interactions and physical propagation processes \citep{han2022predicting}. These methods share with this paper the avoidance of regular grid interpolation, but the computational form of graph message passing and its receptive field expansion differ from Transformer-style global token modeling.

In recent years, neural operators and Transformer-based PDE solvers have further expanded field prediction capabilities on complex geometries and irregular sampling. Geo-FNO \citep{li2023fourier} extends the Fourier operator to complex geometric domains, GNOT \citep{hao2023gnot} learns mappings between input functions, query points, and output fields through geometry-aware Transformers, Transolver \citep{wu2024transolver} uses physics-aware token representations to model complex physical fields, and UPT \citep{alkin2024upt} establishes a more unified token-based representation framework for general physical field prediction. Compared with these methods, APPSolver has a more specific focus: one-step temporal advancement point-wise flow field prediction for ship unstructured CFD meshes, leveraging the spatial organization formed by ship mesh generation strategies and converting non-uniform point density into learnable representation granularity through Adaptive Patch Partitioning.

\section{Method}

\begin{figure}[t]
  \centering
  \includegraphics[width=\textwidth]{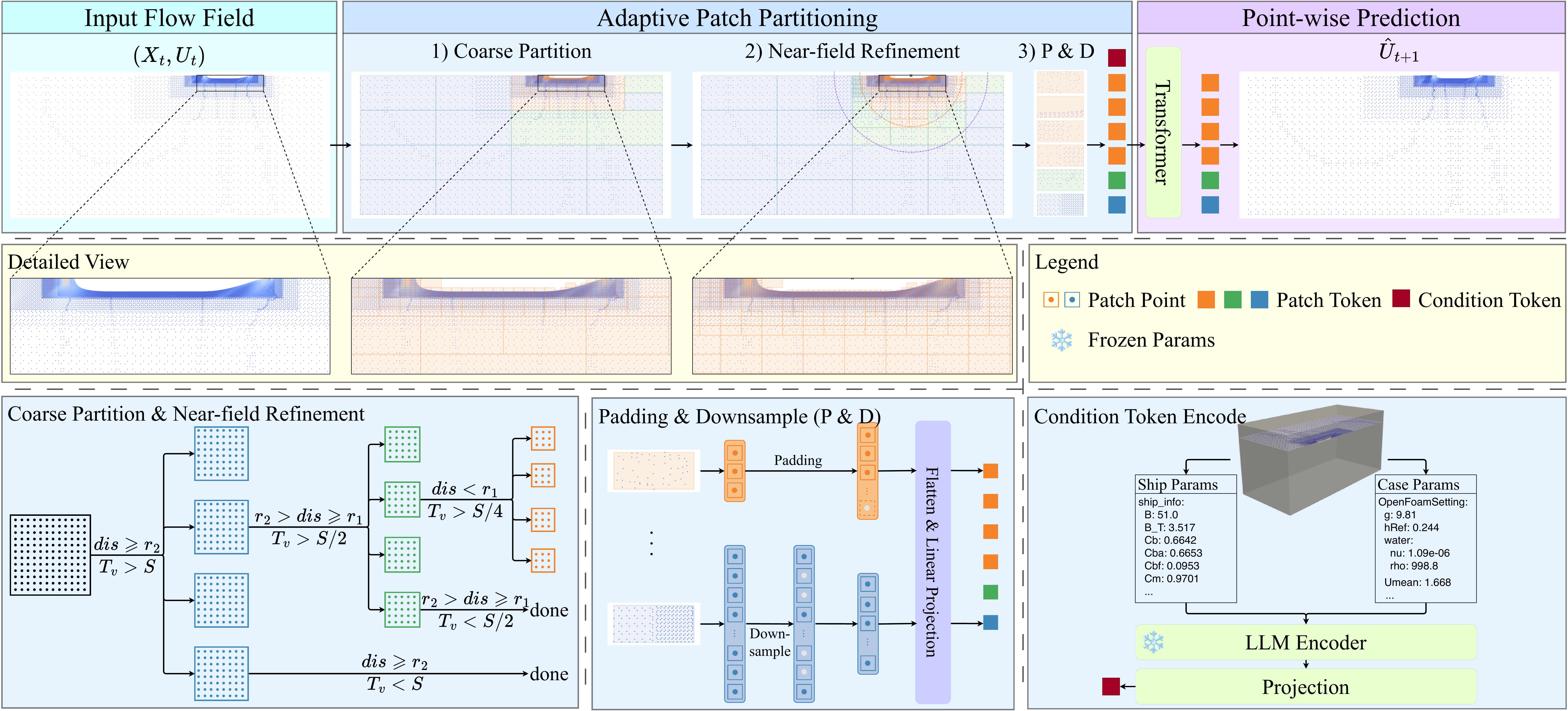}
  \caption{Overview of APPSolver. First, the non-uniform point set near the free surface and the current flow field state are input, followed by coarse partitioning, near-field refinement, padding, and downsampling through APP, converting the original point set into compact patch tokens. Meanwhile, ship geometric parameters and condition parameters are encoded into condition token through a frozen LLM encoder and projection layers, and jointly input with patch tokens into a Transformer backbone, which ultimately outputs the next step point-wise flow field prediction.}
  \label{fig:apps_arch}
\end{figure}

This section presents the overall method design of APPSolver. Figure~\ref{fig:apps_arch} illustrates the complete pipeline from flow field patch partitioning to next-step point-wise flow field prediction: To avoid terminological confusion, APP in this paper specifically refers to the Adaptive Patch Partitioning method, while APPSolver refers to the complete flow field prediction framework encompassing APP, condition token fusion, and the backbone prediction network.

\subsection{Problem definition}
\label{sec:problem_definition}

We formulate ship flow field prediction as a one-step temporal advancement surrogate modeling task on the original unstructured CFD mesh. Given a time-independent point set $\mathcal{X}=\{x_i\}_{i=1}^{N}$ with $x_i\in\mathbb{R}^{d_x}$, the flow field state at time $t$ is $\mathbf{U}_{t}=\{\mathbf{u}_{i,t}\}_{i=1}^{N}$ with $\mathbf{u}_{i,t}\in\mathbb{R}^{d_u}$, where $N$ is the number of mesh points, $d_x=2$ is the coordinate dimensionality, and $d_u=4$ comprises three velocity components and pressure. The model learns a mapping $\mathcal{F}_{\theta}$ conditioned on a case token $c$ that encodes ship geometry and condition parameters:
\begin{equation}
    \widehat{\mathbf{U}}_{t+1}
    =
    \mathcal{F}_{\theta}
    \left(
        \mathcal{X},\;
        \mathbf{U}_{t},\;
        c
    \right),
    \qquad
    \widehat{\mathbf{U}}_{t+1}\in\mathbb{R}^{N\times d_u},
\end{equation}
trained with a point-wise MSE loss $\mathcal{L}(\theta) = |\mathcal{D}|^{-1}\sum_{t}\|\widehat{\mathbf{U}}_{t+1}-\mathbf{U}_{t+1}\|_2^2 / N$. APP serves as an intermediate tokenization mechanism and does not alter the output boundary: the final prediction is reassembled to every original CFD point. A detailed problem formulation is provided in Appendix~\ref{app:problem_definition}.

\subsection{Ship still water navigation dataset}

To construct irregular point set temporal data for ship still water navigation problems, we establish the ship still water navigation dataset (we call it ShipBench) based on the DTC hull from OpenFOAM tutorials\footnote{\url{https://github.com/OpenFOAM/OpenFOAM-7/tree/master/tutorials/multiphase/interFoam/RAS/DTCHullWave}}, as well as KCS and KVLCC2 published by SIMMAN 2020\footnote{\url{https://simman2020.kr/contents/Ship_data_summary.php}}, covering three typical ship types. The data generation pipeline uses the \texttt{interFoam} solver in OpenFOAM \citep{jasak2009openfoam}, combined with the VOF method to capture free surface evolution, with turbulence closure using the SST $k$--$\omega$ model. The geometry and mesh stage uses \texttt{snappyHexMesh} to generate unstructured meshes near the hull, with the initial phase field specified by \texttt{setFields}. Subsequently, unsteady flow fields are obtained through parallel solving, and the point set sequences near the free surface required for learning are exported during post-processing.

The learning target of this dataset focuses on non-uniform two-dimensional point sets near the free surface, rather than full-domain three-dimensional fields on regular grids. Each point is characterized by two-dimensional spatial coordinates, with its physical state comprising three-dimensional velocity components and pressure correction, totaling four channels. To preserve the original non-uniform structure of free surface sampling, this paper does not forcibly resample data onto regular grids or adopt image-like interpolation representations.

Since the number of points and point ordering in the free surface exported at different time steps may differ, we introduc KD-Tree-based alignment during the preprocessing stage to establish point-wise supervision relationships between adjacent time steps. Specifically, the first $N$ coordinates from the first time step are used as reference, where $N$ is the minimum point count across all time steps of the case. Then, for each time step, the point set is reordered according to the nearest-neighbor index based on the reference coordinates and truncated to a uniform length. In practice, the free-surface points are extracted by slicing the waterplane at each time step on the same mesh, so the point locations are inherently consistent across time steps and the alignment procedure does not introduce additional approximation errors. This process is only used for point order alignment during preprocessing and is not an internal model module. After alignment, adjacent time steps are organized into one-step prediction samples. The normalization of physical channels uses Z-score normalization estimated solely from training set statistics to avoid test information leakage.

\subsection{Adaptive patch partitioning}
\label{sec:adaptive_patch_partitioning}

\paragraph{Adaptive flow field partitioning method.}
For the next-step flow field prediction task on the original unstructured CFD mesh, APPSolver first converts the non-uniform point set into compact patch token representations processable by a standard Transformer, using APP as an intermediate tokenization mechanism between the input and the model. Given the two-dimensional spatial coordinates $\mathbf{x}_i=(x_i,y_i)$ at the current time step and their corresponding flow field variables, APP constructs a quadtree partition on the original CFD point set. During the partitioning process, a node is subdivided into four quadrants, and its non-empty child nodes are recursively processed. If subdivision yields only one non-empty quadrant, further partitioning is stopped to avoid unstable recursion in highly non-uniform or locally degenerate cases. The basic stopping criterion for a node can be written as
\begin{equation}
	\mathrm{Stop}(v) = 
	\mathbb{I}\!\left[
		|{\mathcal P}_v| \leq T_v
	\right]
\end{equation}
where ${\mathcal P}_v$ denotes the set of CFD points within node $v$, and $T_v$ denotes the target mesh node capacity for the sub-region corresponding to node $v$. This constraint controls the spatial coverage of different tokens during the recursive partitioning process, allowing APP to form a computable intermediate representation while preserving local structure. The resulting set of leaf nodes constitutes the patch set, where each patch corresponds to a local spatial region in the original unstructured mesh. Within each patch, points are sorted in ascending lexicographic order based on their $(x,y)$ coordinates, i.e., points with smaller $x$-coordinates are placed first, and ties are resolved by increasing $y$-coordinates. This process exploits the spatial organization prior of ship CFD meshes, which are denser near the hull, wake, and free surface and sparser in the far field, while avoiding the high computational and memory overhead of directly using each grid point as a Transformer token.

\paragraph{Near-field refinement strategy.}
During the quadtree partitioning process of APP, the hull proximity, wake region, and free surface typically contain stronger geometric variations and flow field gradients, and therefore require finer-grained patches to represent local structures. Conversely, regions far from the main body perturbation have smoother variations and can use coarser-grained patches to reduce redundant tokens. Inspired by the Signed Distance Function \citep{guo2016convolutional}, we use the distance from the quadtree node center $\mathbf{c}_v$ to the ship center reference point $\mathbf{x}_{\mathrm{ref}}$ as the basis for regional partitioning, normalizing the distance thresholds by ship length scale:
\begin{equation}
r_1=\tau_1 L_{\mathrm{ship}}, \qquad r_2=\tau_2 L_{\mathrm{ship}} .
\end{equation}
Given a base capacity $S$, the target capacity for different distance regions is defined as
\begin{equation}
S_v^{\star} =
\left\{
\begin{array}{ll}
\max(4,\lfloor S/4\rfloor), & \|\mathbf{c}_v-\mathbf{x}_{\mathrm{ref}}\|_2 < r_1, \\[2pt]
\max(4,\lfloor S/2\rfloor), & r_1 \le \|\mathbf{c}_v-\mathbf{x}_{\mathrm{ref}}\|_2 < r_2, \\[2pt]
S, & \|\mathbf{c}_v-\mathbf{x}_{\mathrm{ref}}\|_2 \ge r_2 .
\end{array}
\right.
\end{equation}
The value 4 is the specified minimum patch capacity. Smaller target capacities encourage corresponding nodes to continue subdividing in the quadtree, thereby forming higher patch density in the near field while retaining larger aggregation units in the far field. This strategy does not alter the physical field itself, nor does it interpolate unstructured point clouds onto regular grids. It only adjusts the local token granularity on the original unstructured point set, providing a hierarchical representation that better conforms to the flow physics structure for subsequent spatial downsampling.

\paragraph{Spatial downsampling strategy.}
After quadtree partitioning and near-field refinement, the number of points within different patches is typically inconsistent. To construct patch tokens of uniform length, zero-padding is needed for each patch. However, if the maximum point count after partitioning is directly used as the point slot length, the many points in far-field coarse-grained patches would introduce additional redundancy. To address this, we propose a spatial downsampling strategy. Given a base patch capacity $S$ and a downsampling ratio $\rho$, the target point budget is defined as
\begin{equation}
N_{\mathrm{tar}}=\max(4,\lfloor S\rho\rfloor).
\end{equation}
Near-field patches whose distance to the reference point is less than $r_1$ are preferentially preserved during downsampling. For non-near-field patches whose point count exceeds the budget, approximate uniform sampling is performed along their internal point sequence to bring their size close to $N_{\mathrm{tar}}$. Subsequently, the unified point slot length is determined based on the maximum point count after downsampling, and masks are used to distinguish real points from padding positions. Finally, the input, output, and validity masks are organized as
\begin{equation}
	\mathbf{X}\in\mathbb{R}^{P\times (N_{\mathrm{slot}} C_{\mathrm{in}})},\quad
	\mathbf{Y}\in\mathbb{R}^{P\times (N_{\mathrm{slot}} C_{\mathrm{out}})},\quad
	\mathbf{M}\in\{0,1\}^{P\times N_{mathrm{slot}}}.
\end{equation}
where $P$ is the number of patches and $N_{\mathrm{slot}}$ is the maximum point slot count after downsampling for alignment. This sequence first compresses far-field over-budget patches, then performs fixed-length padding, thereby reducing redundant point slots while maintaining near-field representation granularity. To recover predictions back to all original CFD points, we apply $k$-nearest neighbor interpolation ($k{=}4$) from the downsampled points to the full point set, using distance-weighted averaging based on the original mesh coordinates.

\subsection{Ship geometry and case parameter embedding}

After APP partitions the flow field into patch tokens, APPSolver injects ship geometry and condition parameters into the same token sequence using a prefix token approach: structured parameters are encoded as condition vectors and concatenated with patch tokens in the same representation space.

For the $m$-th sample, structured parameters are denoted as
\begin{equation}
    c^{(m)}=\{c_{\mathrm{geo}}^{(m)},c_{\mathrm{case}}^{(m)}\},
\end{equation}
where $c_{\mathrm{geo}}^{(m)}$ encodes hull geometry (principal dimensions, waterplane parameters, station profiles; see \citep{yan2023influence} for definitions) and $c_{\mathrm{case}}^{(m)}$ encodes operating conditions (speed, Reynolds number, fluid properties, boundary conditions). These parameters are textualized and encoded by a frozen language model \texttt{Qwen3.5-0.8B} \citep{qwen3.5}. This design choice prioritizes simplicity and generality: rather than designing task-specific MLP architectures or hand-crafted Fourier features for different parameter types, a frozen LLM provides a unified, off-the-shelf solution that directly processes textualized descriptions without additional feature engineering:
\begin{equation}
    z_c^{(m)}
    =
    f_{\mathrm{LLM}}\!\left(\mathcal{T}(c^{(m)})\right)_{0}^{\mathrm{last}}
    \in \mathbb{R}^{d_L},
\end{equation}
where $d_L$ is the LLM hidden dimension. A trainable linear layer projects this to the backbone dimension $d_{\mathrm{model}}$:
\begin{equation}
    e_c^{(m)}
    =
	\mathrm{Linear}\!\left(z_c^{(m)}\right)
    \in \mathbb{R}^{d_{\mathrm{model}}}.
\end{equation}
This decouples LLM feature extraction from task-specific dimension adaptation.

Given $P$ patch inputs $\{x_i^{(m)}\}_{i=1}^{P}$ from APP, each patch is projected and augmented with learnable positional encodings:
\[
    p_i^{(m)} = W_x x_i^{(m)} + r_i,\quad i=1,\ldots,P .
\]
The condition token is concatenated as a prefix before patch tokens:
\begin{equation}
    X_0^{(m)}
    =
    \left[
    e_c^{(m)},\;
    p_1^{(m)},\;
    p_2^{(m)},\ldots,
    p_P^{(m)}
    \right]
    \in\mathbb{R}^{(P+1)\times d_{\mathrm{model}}}.
\end{equation}
Both jointly participate in Transformer feature interaction. Only patch representations undergo output projection; the condition token serves as a context modulation signal and is not a supervision target.

\section{Experiments}
\subsection{Experimental setup}
\label{sec:exp_setup}

We evaluate on two datasets across seven evaluation groups. The self-built ShipBench covers DTC, KCS, and KVLCC2 with two speed conditions per hull, and CFDBench \citep{luo2023cfdbench} covers four classical CFD problems: cavity, tube, dam, and cylinder. Our APP-based Transformer and DPT \citep{ranftl2021vision} are compared against six baselines for irregular mesh learning: Transolver \citep{wu2024transolver}, FNO \citep{li2020fno}, UPT \citep{alkin2024upt}, GNOT \cite{hao2023gnot}, Fusion-DeepONet \citep{peyvan2025fusion}, and PCNO \citep{zeng2025point}. All models are configured with approximately 1M parameters. Within each condition, natural adjacent frames are paired into one-step transitions before splitting into training and validation sets, and channel-wise normalization statistics are computed solely from the training split. At evaluation, predictions are assessed over the full reference point set. APP predictions are evaluated after $k$-NN recovery. All models are trained for $16{,}000$ steps with validation every $2{,}000$ steps across model random seeds $42$, $43$, and $44$, and checkpoints are selected by the lowest validation MAE. We report mean MAE, MSE, and RMSE over the three seeds. Detailed dataset construction, training configuration, and hardware setup are provided in Appendix~\ref{app:exp_setup}.

\subsection{Overall comparison}
\label{exp1}

This experiment evaluates APPSolver across different ship hulls and fluid mechanics benchmarks. We conduct a horizontal comparison across seven evaluation groups: S1-DTC, S2-KCS, S3-KVLCC2, C1-Cavity, C2-Tube, C3-Dam, and C4-Cylinder. Each model is trained and validated on all sub-conditions within the corresponding group. The results are reported in Table~\ref{tab:exp1_overall_best_metrics}.

\begin{table}[t]
\centering
\scriptsize
\setlength{\tabcolsep}{2.5pt}
\renewcommand{\arraystretch}{0.82}
\caption{Performance comparison across models on seven groups. Entries show mean MAE/MSE/RMSE over model seeds 42, 43, and 44; the best value for each metric is bolded.}
\label{tab:exp1_overall_best_metrics}
\begin{tabular}{l ccc ccc c}
\toprule
& S1-DTC & S2-KCS & S3-KVLCC2 & C1-Cav & C2-Tube & C3-Dam & C4-Cyl \\
\midrule
Transolver     & .045/.032/.180 & .089/.049/.222 & \textbf{.035}/1.593/1.262 & .012/.001/.028 & .022/\textbf{.006}/\textbf{.075} & .030/.024/.155 & .019/.005/.068 \\
FNO            & .167/.109/.330 & .233/.163/.403 & .082/1.631/1.277 & .078/.028/.168 & .180/.166/.407 & .339/.372/.610 & .245/.236/.486 \\
Fusion-DeepONet& .124/.070/.265 & .177/.105/.325 & .080/1.766/1.329 & .119/.057/.237 & .316/.387/.618 & .122/.099/.315 & .176/.111/.333 \\
UPT            & \textbf{.044}/.034/.184 & .089/.050/.224 & .039/1.640/1.281 & .008/.001/.026 & \textbf{.020}/.009/.094 & .034/.029/.171 & .020/.005/.071 \\
GNOT           & .149/.096/.309 & .204/.138/.371 & .070/1.746/1.321 & .094/.035/.185 & .254/.295/.543 & .148/.131/.361 & .166/.123/.351 \\
PCNO           & .050/\textbf{.032}/\textbf{.179} & \textbf{.082}/\textbf{.042}/\textbf{.206} & .038/\textbf{1.507}/\textbf{1.228} & \textbf{.007}/\textbf{.001}/\textbf{.023} & .065/.014/.119 & \textbf{.022}/\textbf{.007}/\textbf{.085} & \textbf{.012}/\textbf{.001}/\textbf{.031} \\
\midrule
APP-DPT        & .167/.106/.326 & .217/.144/.380 & .055/1.739/1.319 & .122/.071/.267 & .126/.091/.302 & .163/.116/.341 & .188/.127/.357 \\
APP-Transformer& .093/.050/.223 & .129/.067/.259 & .046/1.696/1.302 & .057/.042/.205 & .049/.017/.130 & .075/.058/.240 & .081/.027/.165 \\
\bottomrule
\end{tabular}
\end{table}

\begin{figure}[t]
  \centering
  \includegraphics[width=\textwidth]{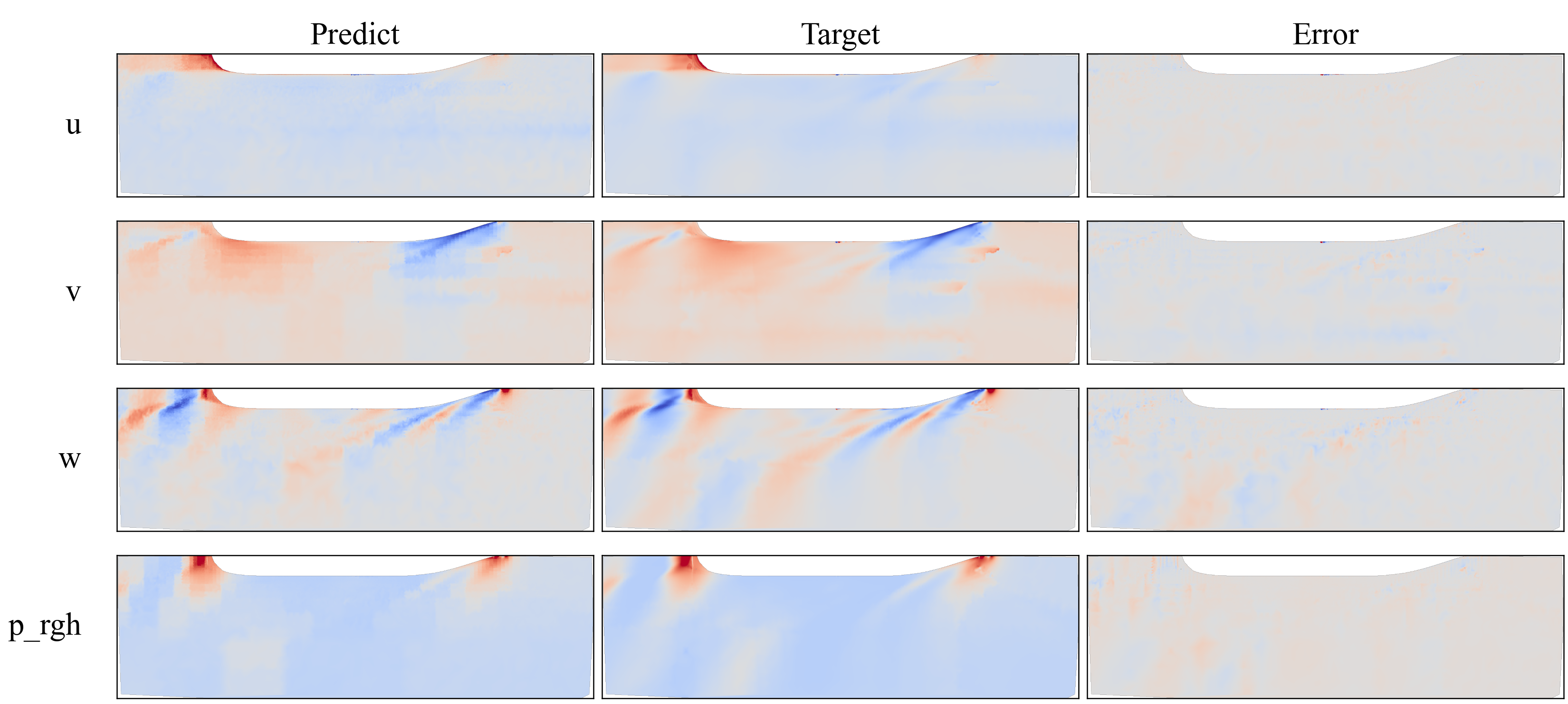}
  \caption{Qualitative next-step point-wise flow field prediction of APP-Transformer on a DTC 1Re sample. Each row corresponds to one physical channel, and only the near-hull region where the mesh is locally refined is shown. Predict, Target, and Error share the per-channel color scale. The model inputs and outputs are point-wise fields, and interpolated fields are shown only for visual presentation.}
  \label{fig:qualitative_app_transformer_uvwp}
\end{figure}

Figure~\ref{fig:qualitative_app_transformer_uvwp} shows, using a time step from the DTC 1Re condition as an example, the near-hull point-wise prediction, target field, and error distribution of APP-Transformer across four flow field channels.

As shown in Table~\ref{tab:exp1_overall_best_metrics}, point-level neural operators achieve the lowest prediction errors across the seven groups. UPT attains the lowest MAE on S1-DTC and C2-Tube. PCNO attains the lowest MAE on S2-KCS, C1-Cavity, C3-Dam, and C4-Cylinder, as well as the lowest MSE and RMSE on S1-DTC and S3-KVLCC2. Transolver attains the lowest MAE on S3-KVLCC2 and the lowest MSE and RMSE on C2-Tube. Across all seven groups, APP-Transformer yields higher prediction errors than the strongest learned baseline in each group, while consistently outperforming APP-DPT across all metrics and achieving lower MAE than FNO, Fusion-DeepONet, and GNOT in every ShipBench and CFDBench group. Combined with the forward computation profiling in Appendix~\ref{app:forward_profiling}, where APP-Transformer requires $1.815$ GFLOPs on a representative DTC sample compared with $1.694$ GFLOPs for PCNO and $7.482$ to $53.644$ GFLOPs for the remaining point-level baselines, these results characterize an accuracy--efficiency trade-off between patch-token compression and point-level resolution.

\subsection{Adaptive partition and learned-token ablations}

We compare the production APP-Transformer across three partition strategies under the same $16{,}000$-step training protocol over seeds $42$, $43$, and $44$. Adaptive uses the proposed distance-refined quadtree. Uniform disables distance refinement while retaining the remaining APP pipeline. Uniform + learned augments the uniform partition with a Transolver-style 32-slice soft assignment inside each backbone block. Table~\ref{tab:patch_strategy} shows that adaptive partitioning achieves the lowest error on all three hulls, reducing MAE relative to uniform + learned by $24.9\%$, $19.7\%$, and $16.4\%$ on DTC, KCS, and KVLCC2, respectively. These results support the distance-refined spatial prior in APPSolver.

\begin{table}[t]
\centering
\normalsize
\caption{Partition strategy ablation for the production APP-Transformer on ShipBench. Entries report normalized MAE as mean $\pm$ sample standard deviation over seeds 42, 43, and 44; the best value for each hull is bolded.}
\label{tab:patch_strategy}
\begin{tabular}{lccc}
\toprule
Strategy & DTC & KCS & KVLCC2 \\
\midrule
Adaptive & \textbf{.09254$\pm$.00016} & \textbf{.12910$\pm$.00125} & \textbf{.04600$\pm$.00085} \\
Uniform & .12431$\pm$.00152 & .16564$\pm$.00085 & .04890$\pm$.00184 \\
Uniform + learned & .12324$\pm$.00026 & .16086$\pm$.00022 & .05502$\pm$.00046 \\
\bottomrule
\end{tabular}
\end{table}

\subsection{APP ablation on patch size and downsample ratio}

This subsection evaluates the compression and recovery characteristics of the deterministic APP representation across different patch capacities and downsample ratios. Because APP maps downsampled patch points back to the original CFD point set via $k$-NN interpolation at the output stage, the recovery accuracy of this representation directly establishes the baseline fidelity of the tokenized point set. On the ShipBench point sets, we evaluate base patch capacities $S\in\{256,384,512,768,1024\}$ and downsample ratios $\rho\in\{0.60,0.70,0.80,0.90,1.00\}$ to quantify the compression ratios and point-wise recovery errors produced by the APP representation.

Let $n_p^{\mathrm{orig}}$ and $n_p^{\mathrm{samp}}$ denote the number of valid points in patch $p$ before and after APP downsampling, respectively. Let $K^{\mathrm{orig}}=\max_p n_p^{\mathrm{orig}}$ and $K^{\mathrm{samp}}=\max_p n_p^{\mathrm{samp}}$ be the corresponding padded point-slot lengths. For a sample with $P$ patches, the padded storage sizes before and after downsampling are defined as $M_{\mathrm{before}}=P K^{\mathrm{orig}}$ and $M_{\mathrm{after}}=P K^{\mathrm{samp}}$, where the channel dimension is omitted because it is shared by both terms. We further define $\bar{n}_{\mathrm{original}}=P^{-1}\sum_{p=1}^{P} n_p^{\mathrm{orig}}$ and $\bar{n}_{\mathrm{sampled}}=P^{-1}\sum_{p=1}^{P} n_p^{\mathrm{samp}}$. The padding compression ratio is defined as $\mathrm{PCR} = M_{\mathrm{before}} / M_{\mathrm{after}}$ and measures the reduction in padded tensor slots. The effective data compression rate is defined as $\mathrm{EDCR} = 1 - \bar{n}_{\mathrm{sampled}} / \bar{n}_{\mathrm{original}}$ and measures the actual reduction of valid points after downsampling. Recovery error is measured by relative RMSE, denoted as Rel.\ RMSE, after reconstructing the original point set from the sampled points.

\begin{table}[t]
\centering
\normalsize
\caption{APP representation ablation on base patch capacity $S$ and downsample ratio $\rho$, reporting PCR, EDCR, and recovery Rel.\ RMSE.}
\label{tab:exp2_ablation}
\begin{minipage}[t]{0.5\textwidth}
\vspace{0pt}
\centering
\setlength{\tabcolsep}{3pt}
\begin{tabular}{cc ccc}
\toprule
$S$ & $\rho$ & PCR & EDCR [\%] & Rel.\ RMSE \\
\midrule
256 & 0.60 & 3.159 & 4.579 & 0.1278 \\
256 & 0.70 & 2.700 & 2.559 & 0.0862 \\
256 & 0.80 & 2.369 & 1.323 & 0.0453 \\
256 & 0.90 & 2.104 & 0.683 & 0.0171 \\
256 & 1.00 & 1.926 & 0.534 & 0.0143 \\
\midrule
384 & 0.60 & 2.501 & 11.863 & 0.1350 \\
384 & 0.70 & 2.147 & 7.903 & 0.0929 \\
384 & 0.80 & 1.874 & 4.286 & 0.0611 \\
384 & 0.90 & 1.668 & 1.478 & 0.0311 \\
384 & 1.00 & 1.506 & 0.438 & 0.0132 \\
\midrule
512 & 0.60 & 2.162 & 15.501 & 0.1772 \\
512 & 0.70 & 1.854 & 7.346 & 0.1337 \\
512 & 0.80 & 1.623 & 2.453 & 0.0871 \\
512 & 0.90 & 1.443 & 0.780 & 0.0390 \\
512 & 1.00 & 1.296 & 0.342 & 0.0148 \\
\bottomrule
\end{tabular}
\end{minipage}%
\begin{minipage}[t]{0.5\textwidth}
\vspace{0pt}
\centering
\setlength{\tabcolsep}{3pt}
\begin{tabular}{cc ccc}
\toprule
$S$ & $\rho$ & PCR & EDCR [\%] & Rel.\ RMSE \\
\midrule
768 & 0.60 & 1.693 & 4.524 & 0.1032 \\
768 & 0.70 & 1.451 & 2.089 & 0.0559 \\
768 & 0.80 & 1.269 & 0.900 & 0.0210 \\
768 & 0.90 & 1.160 & 0.293 & 0.0135 \\
768 & 1.00 & 1.086 & 0.150 & 0.0106 \\
\midrule
1024 & 0.60 & 1.575 & 6.817 & 0.1551 \\
1024 & 0.70 & 1.351 & 4.069 & 0.1160 \\
1024 & 0.80 & 1.181 & 1.932 & 0.0742 \\
1024 & 0.90 & 1.059 & 0.431 & 0.0228 \\
1024 & 1.00 & 1.000 & 0.000 & 0.0000 \\
\bottomrule
\end{tabular}
\end{minipage}
\end{table}

Table~\ref{tab:exp2_ablation} summarizes the representation-level compression metrics and recovery errors across all 25 configurations. PCR and recovery Rel.\ RMSE exhibit a direct trade-off: as the base patch capacity $S$ or the downsample ratio $\rho$ increases, more points are retained and recovery error decreases alongside lower compression. For $S=256$, increasing $\rho$ from $0.60$ to $1.00$ lowers Rel.\ RMSE from $0.1278$ to $0.0143$ while reducing PCR from $3.159$ to $1.926$ and EDCR from $4.579\%$ to $0.534\%$. Under larger patch capacities, such as $S=768$ with $\rho=1.00$, Rel.\ RMSE further decreases to $0.0106$, while compression drops to $\mathrm{PCR}=1.086$ and $\mathrm{EDCR}=0.150\%$. In our experiments, we use $S=256$ and $\rho=0.60$ as the default compact configuration.

\subsection{Role of condition token in cross-hull generalization}

We evaluate the condition interface on APP-Transformer by comparing zero conditioning, a normalized numeric MLP, a Fourier-feature MLP, FiLM, Qwen2.5~\citep{qwen2.5}, and Qwen3.5~\citep{qwen3.5} across the full ShipBench dataset and three leave-one-hull-out scenarios. Table~\ref{tab:condition_encoders} reports the three-seed mean normalized MAE. Under the full ShipBench setting, zero conditioning achieves the lowest MAE. Across all three leave-one-hull-out settings, Qwen3.5 achieves the lowest MAE, reducing error relative to zero conditioning by $3.0\%$ on DTC, $4.5\%$ on KCS, and $1.0\%$ on KVLCC2. Numeric encoders remain competitive, as the numeric MLP reduces MAE by $2.5\%$ on held-out DTC, while Qwen2.5 and the Fourier MLP reduce MAE by $3.0\%$ and $2.8\%$ on held-out KCS. When KVLCC2 is held out, all methods incur large absolute errors because its hull geometry differs substantially from the training hulls DTC and KCS. Under this distribution shift, the Qwen3.5 condition token still reduces MAE by $1.0\%$, indicating that condition-token fusion provides modest benefit for generalization. This incremental gain does not constitute emergent generalization.

\begin{table}[t]
\centering
\normalsize
\setlength{\tabcolsep}{4.5pt}
\caption{Condition encoder comparison on APP-Transformer across full ShipBench and leave-one-hull-out settings. Entries report mean normalized MAE over seeds 42, 43, and 44; the best value in each row is bolded.}
\label{tab:condition_encoders}
\begin{tabular}{lcccccc}
\toprule
Scenario & Zero & Numeric MLP & Fourier MLP & FiLM & Qwen2.5 & Qwen3.5 \\
\midrule
Full ShipBench & \textbf{.03979} & .04048 & .04133 & .04036 & .04069 & .04128 \\
Leave out DTC & .02690 & .02623 & .02859 & .17942 & .02717 & \textbf{.02609} \\
Leave out KCS & .04395 & .04779 & .04273 & .23264 & .04262 & \textbf{.04199} \\
Leave out KVLCC2 & 7.84960 & 8.24206 & 8.00103 & 8.50888 & 7.89234 & \textbf{7.77022} \\
\bottomrule
\end{tabular}
\end{table}

\section{Limitations and conclusion}

This paper still has several limitations. First, the current algorithm uses two-dimensional free-surface coordinates with $d_x=2$ as geometric input, and has not yet validated three-dimensional coordinate inputs or volume mesh modeling. Second, the Ship Still Water Navigation Dataset/ShipBench is still relatively limited in scale. Further expansion of hull types, speeds, and operating condition ranges is needed to test generalization capability under larger parameter models and wider distributions. Finally, the main evaluation focuses on one-step point-wise flow field prediction, while the $50$-step autoregressive rollout comparison in Appendix~\ref{app:rollout} shows that all learned models accumulate substantial error over long horizons and yield higher cumulative MAE than the persistence baseline.

In summary, this paper proposes APPSolver, formulating ship unstructured CFD mesh flow field prediction as a one-step temporal advancement point-wise flow field prediction problem. Its core mechanism, Adaptive Patch Partitioning, organizes non-uniform free surface points into patch tokens, reducing token scale while preserving the spatial organization of ship CFD meshes. Experimental results show that APP-Transformer exhibits an accuracy--efficiency trade-off, yielding higher prediction errors than the strongest learned baselines across ShipBench and CFDBench, while forward profiling on a representative DTC input shows substantially lower computation than most point-level and attention-based baselines. Ablation studies validate the distance-refined spatial prior and characterize the trade-off between compression and recovery fidelity. Cross-hull generalization experiments indicate that condition gains vary across held-out hulls and encoder choices.

\begin{ack}
This work was supported by the National Key R\&D Program of China under Grant No. 2022YFB3306200. The authors declare no competing interests.
\end{ack}

{\small
\bibliographystyle{plainnat}
\bibliography{reference}
}

\appendix

\section{Detailed experimental setup}
\label{app:exp_setup}

This appendix provides the full experimental configuration described in Section~\ref{sec:exp_setup}.

\subsection{Datasets}

We evaluate model performance on two data domains: the self-built Ship Still Water Navigation Dataset, referred to as ShipBench, and CFDBench \citep{luo2023cfdbench}. ShipBench covers three typical hull types: DTC, KCS, and KVLCC2, each containing two operating conditions: 1$\times$ design speed, denoted as 1Re, and 2$\times$ design speed, denoted as 2Re, with model ship lengths of approximately $6$--$8\,\mathrm{m}$. Each case contains approximately $70\,\mathrm{s}$ of unsteady evolution, with an export interval of $0.2\,\mathrm{s}$, corresponding to approximately $350$ time steps. The number of points near the free surface is approximately $26\mathrm{k}$--$44\mathrm{k}$. In this data domain, the task is set as one-step prediction from the current time step to the next time step, using two-dimensional coordinates and current physical states on the irregular free surface point set to predict three-dimensional velocity components and pressure correction at the next time step. For CFDBench, we use its original data rather than interpolated image-like data, and process it as irregular point set time series following the same approach as ShipBench, selecting four classic CFD problems: cavity, tube, dam, and cylinder, with samples covering different boundary conditions, physical parameters, and computational domain geometries.

\subsection{Baselines and evaluation protocol}

Comparison methods include our APP-based Transformer and DPT, as well as baseline methods for irregular mesh or point set learning: Transolver \citep{wu2024transolver}, FNO \citep{li2020fno}, UPT \citep{alkin2024upt}, GNOT \cite{hao2023gnot}, Fusion-DeepONet \citep{peyvan2025fusion}, and PCNO \citep{zeng2025point}. The first four baselines come from Neural-Solver-Library\footnote{\url{https://github.com/thuml/Neural-Solver-Library}} and are run under unified data interfaces, training pipelines, and evaluation protocols. We adjust model parameter configurations so that all models have approximately 1M parameters, thereby ensuring that comparisons reflect differences in model representation and solving mechanisms rather than implementation details.

\subsection{Training configuration}

Training uses a multi-condition joint training setup. Within each condition, temporal samples are split into training and test sets with a $0.8/0.2$ ratio. The main experiment uses a fixed training step scheme, training for $16{,}000$ steps total, with logging and evaluation every $500$ steps, a random seed of $42$, and a batch size of $4$. The default APP patch size is $256$ and the downsample ratio is $0.6$. In Exp2, we further vary the patch size and downsample ratio to analyze the effects of intermediate representation granularity and sampling ratio on prediction error and computational overhead. In Exp3, we separately evaluate the contributions of condition token and LLM embeddings to examine the effect of conditional information injection on cross-condition modeling.

\subsection{Hardware and metrics}

All experiments run on the AutoDL\footnote{\url{https://autodl.com}} platform, with hardware including one NVIDIA GeForce RTX 4080 SUPER GPU with 32 GB memory, dual Intel Xeon Platinum 8352V processors with 128 logical CPUs, and 503 GiB system memory. Evaluation metrics use MAE, MSE, and RMSE. For APP patch models, loss are computed only at valid mask positions, and metrics are computed after k-NN recovery to avoid the impact of padding points on metrics. Under the multi-condition joint training setup, we report both per-group internal metrics and cross-group average metrics to characterize the model's overall performance across different hull types, flow problems, and geometric conditions.

\section{Forward computation profiling}
\label{app:forward_profiling}

Table~\ref{tab:exp1_forward_profile} reports the forward computational cost of each model on a representative DTC sample. This test uses the 1M parameter model configuration and APP settings from the experimental setup. The APP models take $P=1171$ patch tokens as input, with a maximum point slot count of $256$ per patch, while the irregular point set baselines directly use $N=26101$ points as input. APP-Transformer and APP-DPT have significantly lower forward FLOPs than most baselines including Transolver, GNOT, UPT, Fusion-DeepONet, and FNO, indicating that after APP converts large-scale unstructured point sets into compact patch tokens, it can effectively reduce the forward computational overhead of most token/attention-based or point-level representation models. Meanwhile, PCNO has slightly lower FLOPs under profiler statistics, but its CPU forward time is higher, reflecting that different operator implementations, scatter/spectral operators, and hardware backends still impact actual runtime.

\begin{table}[h]
\centering
\normalsize
\setlength{\tabcolsep}{4pt}
\caption{Forward profiling on a representative DTC sample. APP models use $P=1171$ patch tokens with a maximum point slot count of $153$; irregular baselines use $N=26101$ points. FLOPs are measured by \texttt{torch.profiler} at batch size 1; CPU time is single-threaded forward time, provided only as implementation-level reference.}
\label{tab:exp1_forward_profile}
\begin{tabular}{lcccc}
\toprule
Model & Input units & Params [M] & FLOPs [G] & CPU time [ms] \\
\midrule
APP-Transformer & $P=1171$ & 0.930 & 1.815 & \textbf{52.8} \\
APP-DPT & $P=1171$ & 0.939 & 2.047 & 68.7 \\
Transolver & $N=26101$ & 0.989 & 53.644 & 319.7 \\
FNO & $N=26101$ & 0.990 & 7.482 & 238.9 \\
Fusion-DeepONet & $N=26101$ & 0.997 & 23.598 & 315.0 \\
UPT & $N=26101$ & 1.000 & 18.936 & 155.5 \\
GNOT & $N=26101$ & 0.993 & 52.705 & 571.5 \\
PCNO & $N=26101$ & 1.033 & \textbf{1.694} & 265.4 \\
\bottomrule
\end{tabular}
\end{table}

\section{50-step autoregressive rollout}
\label{app:rollout}

To evaluate multi-step behavior beyond single-step prediction, we test all eight learned models alongside the deterministic persistence baseline $\widehat{\mathbf{U}}_{t+1}=\mathbf{U}_{t}$ on ShipBench. For each learned model, we use the seed 42 checkpoint with the lowest validation MAE. Each operating condition reserves an independent continuous sequence of $51$ frames, and each model recursively feeds its predicted state back as input for $50$ rollout steps. Table~\ref{tab:rollout_all_models} reports the cumulative MAE ratio through horizon $h=50$, defined as cumulative model MAE divided by cumulative persistence MAE over the $50$-step horizon.

\begin{table}[h]
\centering
\normalsize
\caption{Autoregressive 50-step rollout comparison on ShipBench using seed 42 best-validation-MAE checkpoints. Entries report the cumulative MAE ratio through $h=50$, defined as cumulative model MAE divided by cumulative persistence MAE; the lowest ratio for each hull is bolded.}
\label{tab:rollout_all_models}
\begin{tabular}{lccc}
\toprule
Model & DTC & KCS & KVLCC2 \\
\midrule
APP-Transformer & 6.01 & 3.81 & 6.44 \\
APP-DPT & 4.78 & 5.54 & 51.49 \\
Transolver & 3.86 & 5.82 & 40.55 \\
UPT & 2.77 & 3.07 & 22.93 \\
GNOT & 1.70 & 1.98 & 10.14 \\
FNO & 8.20 & 6.91 & 7.05 \\
Fusion-DeepONet & 4.69 & 6.10 & 9.87 \\
PCNO & 4.63 & 8.78 & Non-finite at $h=19$ \\
\midrule
Persistence & \textbf{1.00} & \textbf{1.00} & \textbf{1.00} \\
\bottomrule
\end{tabular}
\end{table}

As shown in Table~\ref{tab:rollout_all_models}, the persistence baseline achieves the lowest cumulative MAE across all three hulls. Among the learned models, GNOT obtains the lowest cumulative MAE ratios on DTC and KCS at $1.70$ and $1.98$, respectively, whereas APP-Transformer yields cumulative MAE ratios of $6.01$ on DTC, $3.81$ on KCS, and $6.44$ on KVLCC2. PCNO produces non-finite predictions at step $h=19$ on KVLCC2. Across all learned models, training solely with the one-step absolute next-state objective leads to substantial error accumulation during long-horizon autoregressive rollout.

\section{Detailed problem formulation}
\label{app:problem_definition}

This appendix provides the full formulation of the task defined in Section~\ref{sec:problem_definition}.

For a ship still water navigation CFD case, the preprocessed free surface computational domain can be viewed as a time-independent non-uniform point set
\begin{equation}
    \mathcal{X}=\{x_i\}_{i=1}^{N},
    \qquad
    x_i\in\mathbb{R}^{d_x},
\end{equation}
where $N$ is the number of free surface CFD mesh points for the case and $d_x$ is the coordinate dimension. In the ship still water navigation cases, the $x$ and $y$ coordinate channels of the free surface are selected, so the experimental setup in this paper is $d_x=2$, without considering the $z$-axis coordinate.

At each mesh point, the flow field state is denoted as
\begin{equation}
    \mathbf{U}_{t}
    =
    \{\mathbf{u}_{i,t}\}_{i=1}^{N},
    \qquad
    \mathbf{u}_{i,t}\in\mathbb{R}^{d_u}.
\end{equation}
The current dataset adopts four physical channels: velocity components $u, v, w$ in the $x, y, z$ directions and reference pressure $p$, i.e., $d_u=4$.

Since this paper focuses on unsteady temporal advancement, point-wise supervision relationships need to be established between adjacent time steps. Under the above time-independent mesh definition, for the input point $x_i$ at the current time step, its supervision label is directly taken as the flow field value $\mathbf{u}_{i,t+1}$ at the same spatial location at the next time step. The alignment of point order across time steps in the actual data is accomplished by dataset preprocessing and is described in the next subsection.

Building upon this, this paper learns a parameterized mapping $\mathcal{F}_{\theta}$, taking the time-independent original CFD coordinates, the current time step flow field state, and case conditions as inputs, to predict the point-wise flow field at the next time step:
\begin{equation}
    \widehat{\mathbf{U}}_{t+1}
    =
    \mathcal{F}_{\theta}
    \left(
        \mathcal{X},
        \mathbf{U}_{t},
        c
    \right),
    \qquad
    \widehat{\mathbf{U}}_{t+1}\in\mathbb{R}^{N\times d_u}.
\end{equation}
where $c$ denotes the condition token encoded from the ship geometric parameters and navigation operating condition parameters of the case. The default time interval setting is 1, meaning the task is one-step prediction from $t$ to $t+1$, rather than multi-step rollout prediction or steady-state field reconstruction.

The training objective employs a point-wise supervision loss, minimizing the error between the predicted flow field and the next time-step flow field across all training cases and time steps:
\begin{equation}
    \mathcal{L}(\theta)
    =
    \frac{1}{|\mathcal{D}|}
    \sum_{t\in\mathcal{D}}
    \frac{1}{N}
    \sum_{i=1}^{N}
    \left\|
        \widehat{\mathbf{u}}_{i,t+1}
        -
        \mathbf{u}_{i,t+1}
    \right\|_2^2 .
\end{equation}

It should be emphasized that Adaptive Patch Partitioning (APP) serves as an intermediate representation mechanism in this paper, used to adaptively aggregate large-scale unstructured point sets into patch tokens. Geometric and operating condition information can be encoded as condition token and jointly modeled with patch tokens in a unified Transformer input space. APP does not alter the output boundary of the problem definition: after training a flow field prediction model on data processed by the APP method, the final predictions still need to be reassembled back to every original CFD point, achieving next-step point-wise flow field prediction.

\section{Additional related work}
\label{app:related_work}

\subsection{Data-driven ship hydrodynamics and design surrogates}

Ship hydrodynamic performance evaluation has long relied on model testing, empirical formulas, and high-fidelity CFD \citep{bertram2012practical}. Model testing provides reliable physical observations but is costly and difficult to cover large-scale hull types and operating condition combinations. Empirical and semi-empirical methods offer high engineering efficiency but are typically limited in scope and geometric parameterization. In contrast, CFD methods based on RANS or LES can resolve spatial distribution characteristics of pressure, velocity, free surface, and wake under complex hull geometries, and have become commonly used tools in modern ship design analysis \citep{jasak2009openfoam,larsson2013cfd}. However, when the design process requires repeated evaluation of different hull types, speeds, or attitudes, the computational cost of traditional CFD still constitutes a significant bottleneck.

To reduce simulation costs, researchers have recently begun constructing data-driven surrogate models for ship hydrodynamic problems \citep{kaklis2025machine}. Such methods typically utilize existing CFD or experimental data to learn approximate mappings from hull parameters and operating conditions to resistance, heave, trim, or local flow fields, thereby serving rapid performance evaluation and design space exploration \citep{alexiou2022towards,karagiannidis2021data,gupta2022ship}. These works demonstrate that deep learning models can approximate high-cost numerical solving processes within specific hull families and operating condition ranges. This paper shares the same goal as the aforementioned research, aiming to improve ship flow field prediction efficiency. The difference is that we focus not on overall hydrodynamic coefficients or field quantity regression on regular grids, but on one-step temporal advancement point-wise flow field prediction for ship unstructured CFD meshes.

\subsection{Conditional and token-based representations for physical fields}

Token-based representations have become a common form in visual modeling and physical field learning. The Transformer \citep{vaswani2017attention} originally established global dependencies between sequence tokens through self-attention. ViT \citep{dosovitskiy2020image} divides images into patch tokens, enabling standard Transformers to process two-dimensional visual inputs. DETR \citep{carion2020end} formulates the detection task as a set prediction problem through object queries and cross-attention. These works demonstrate that different types of tokens, such as patches, queries, and conditions, can serve as unified interfaces to jointly incorporate local observations, global context, and task conditions into the same attention modeling framework.

In scientific machine learning, conditional input is also one of the commonly used mechanisms for building generalizable physical field models. Operator learning methods such as DeepONet \citep{lu2019deeponet, lu2021learning} learn the compositional relationship between input functions and query positions through branch--trunk structures, providing a foundational form for subsequent conditional neural operators. Methods such as GNOT, Transolver, and UPT further organize geometry, physical parameters, query points, or field variables into tokens or latent representations, enabling conditional prediction in complex geometries and multi-physics tasks \citep{hao2023gnot,wu2024transolver,alkin2024upt}. Similarly, the design focus of APPSolver is to place the patch tokens obtained from adaptive partitioning of ship unstructured flow fields and condition token within a unified token space, so that the current flow field state, hull geometry, and operating condition information jointly participate in one-step temporal advancement point-wise flow field prediction, while maintaining explicit utilization of ship CFD mesh spatial organization.

\newpage
\input{checklist.tex}

\end{document}

%% file: checklist.tex
\section*{NeurIPS Paper Checklist}

\begin{enumerate}

\item {\bf Claims}
    \item[] Question: Do the main claims made in the abstract and introduction accurately reflect the paper's contributions and scope?
    \item[] Answer: \answerYes{} 
    \item[] Justification: We provide clear claims in the abstract and introduction for our paper's contributions and scope.

\item {\bf Limitations}
    \item[] Question: Does the paper discuss the limitations of the work performed by the authors?
    \item[] Answer: \answerYes{} 
    \item[] Justification: We discuss the limitations of our work in the Limitations and Conclusion section.

\item {\bf Theory assumptions and proofs}
    \item[] Question: For each theoretical result, does the paper provide the full set of assumptions and a complete (and correct) proof?
    \item[] Answer: \answerNA{} 
    \item[] Justification: We do not rely on theoretical assumptions and any proofs in mathematics.

    \item {\bf Experimental result reproducibility}
    \item[] Question: Does the paper fully disclose all the information needed to reproduce the main experimental results of the paper to the extent that it affects the main claims and/or conclusions of the paper (regardless of whether the code and data are provided or not)?
    \item[] Answer: \answerYes{} 
    \item[] Justification: We present the detailed information in Experiments section and Appendix A.

\item {\bf Open access to data and code}
    \item[] Question: Does the paper provide open access to the data and code, with sufficient instructions to faithfully reproduce the main experimental results, as described in supplemental material?
    \item[] Answer: \answerYes{} 
    \item[] Justification: Our source code is available at \url{https://github.com/wenhuahuo/APPSolver} and is included in the supplemental material.

\item {\bf Experimental setting/details}
    \item[] Question: Does the paper specify all the training and test details (e.g., data splits, hyperparameters, how they were chosen, type of optimizer) necessary to understand the results?
    \item[] Answer: \answerYes{} 
    \item[] Justification: We present the detailed information in Experiments section and Appendix A.

\item {\bf Experiment statistical significance}
    \item[] Question: Does the paper report error bars suitably and correctly defined or other appropriate information about the statistical significance of the experiments?
    \item[] Answer: \answerNo{} 
    \item[] Justification: Due to computational constraints, each experiment was conducted once. We report deterministic evaluation metrics.

\item {\bf Experiments compute resources}
    \item[] Question: For each experiment, does the paper provide sufficient information on the computer resources (type of compute workers, memory, time of execution) needed to reproduce the experiments?
    \item[] Answer: \answerYes{} 
    \item[] Justification: We provide the computation details in Appendix A.
    
\item {\bf Code of ethics}
    \item[] Question: Does the research conducted in the paper conform, in every respect, with the NeurIPS Code of Ethics \url{https://neurips.cc/public/EthicsGuidelines}?
    \item[] Answer: \answerYes{} 
    \item[] Justification:  We strictly follow the NeurIPS Code of Ethics.

\item {\bf Broader impacts}
    \item[] Question: Does the paper discuss both potential positive societal impacts and negative societal impacts of the work performed?
    \item[] Answer: \answerNo{} 
    \item[] Justification: Our paper is about foundational methodological research for CFD-based ship hydrodynamics prediction with no direct path to negative societal impacts.
    
\item {\bf Safeguards}
    \item[] Question: Does the paper describe safeguards that have been put in place for responsible release of data or models that have a high risk for misuse (e.g., pre-trained language models, image generators, or scraped datasets)?
    \item[] Answer: \answerNA{} 
    \item[] Justification: Our paper poses no risks of misuse.

\item {\bf Licenses for existing assets}
    \item[] Question: Are the creators or original owners of assets (e.g., code, data, models), used in the paper, properly credited and are the license and terms of use explicitly mentioned and properly respected?
    \item[] Answer: \answerYes{} 
    \item[] Justification: We added necessary citations when we use related public dataset and codes/models for baselines.

\item {\bf New assets}
    \item[] Question: Are new assets introduced in the paper well documented and is the documentation provided alongside the assets?
    \item[] Answer: \answerYes{} 
    \item[] Justification: ShipBench is a newly constructed dataset. We will release the OpenFOAM simulation code and dataset construction guide.

\item {\bf Crowdsourcing and research with human subjects}
    \item[] Question: For crowdsourcing experiments and research with human subjects, does the paper include the full text of instructions given to participants and screenshots, if applicable, as well as details about compensation (if any)? 
    \item[] Answer: \answerNA{} 
    \item[] Justification: Our paper does not involve crowdsourcing nor research with human subjects.

\item {\bf Institutional review board (IRB) approvals or equivalent for research with human subjects}
    \item[] Question: Does the paper describe potential risks incurred by study participants, whether such risks were disclosed to the subjects, and whether Institutional Review Board (IRB) approvals (or an equivalent approval/review based on the requirements of your country or institution) were obtained?
    \item[] Answer: \answerNA{} 
    \item[] Justification: Our paper does not involve crowdsourcing nor research with human subjects.

\item {\bf Declaration of LLM usage}
    \item[] Question: Does the paper describe the usage of LLMs if it is an important, original, or non-standard component of the core methods in this research? Note that if the LLM is used only for writing, editing, or formatting purposes and does \emph{not} impact the core methodology, scientific rigor, or originality of the research, declaration is not required.
    \item[] Answer: \answerYes{} 
    \item[] Justification: We use the frozen language models Qwen2.5-0.5B~\citep{qwen2.5} and Qwen3.5-0.8B~\citep{qwen3.5} as condition encoders that map textualized ship geometry and operating condition descriptions to condition tokens, and we compare them against zero conditioning and lightweight numerical encoders in the Experiments section.

\end{enumerate}